\documentclass[a4paper,twoside]{article}

\usepackage{epsfig}
\usepackage{calc}
\usepackage{amssymb}
\usepackage{amstext}
\usepackage{amsmath}
\usepackage{multicol}
\usepackage{pslatex}
\usepackage{apalike}
\usepackage{algorithm2e}
\usepackage{hyperref}
\usepackage[bottom]{footmisc}
\usepackage{float}
\usepackage{graphicx}
\graphicspath{{Images/}} 
\usepackage{eso-pic} 
\usepackage{subfig} 
\usepackage{caption} 
\usepackage{transparent}
\definecolor{lightgreen}{RGB}{144,238,144}
\definecolor{orange}{RGB}{255,165,0}
\definecolor{yellow}{rgb}{255,255,0}
\definecolor{red}{rgb}{255,0,0}
\usepackage{tabularx}
\usepackage{longtable} 
\usepackage{colortbl}

\newcommand{\free}{\mathcal{M}}  

\newcommand{\norm}[1]{\left\lVert#1\right\rVert}

\usepackage{SCITEPRESS}     

\begin{document}

\title{RoboMorph: In-Context Meta-Learning for Robot Dynamics Modeling}

\author{\authorname{Manuel Bianchi Bazzi\sup{1}, Asad Ali Shahid\sup{2}, Christopher Agia\sup{3}, John Alora\sup{3}, Marco Forgione\sup{2}, Dario Piga\sup{2}, Francesco Braghin\sup{1}, Marco Pavone\sup{3}, Loris Roveda\sup{2,3}}
\affiliation{\sup{1}Department of Mechanical Engineering, Politecnico di Milano, Italy}
\affiliation{\sup{2}Istituto Dalle Molle di studi sull'intelligenza artificiale (IDSIA USI-SUPSI), Scuola universitaria professionale della Svizzera italiana, DTI (email: loris.roveda@idsia.ch)}
\affiliation{\sup{3}Stanford University}
}

\keywords{Transformers, In-context Learning, Meta Learning, Transfer Learning, Deep Learning, Isaac Gym, Robot Dynamics Modeling.}

\abstract{The landscape of Deep Learning has experienced a major shift with the pervasive adoption of Transformer-based architectures, particularly in Natural Language Processing (NLP). Novel avenues for physical applications, such as solving Partial Differential Equations and Image Vision, have been explored. However, in challenging domains like robotics, where high non-linearity poses significant challenges, Transformer-based applications are scarce. While Transformers have been used to provide robots with knowledge about high-level tasks, few efforts have been made to perform system identification.
This paper proposes a novel methodology to learn a meta-dynamical model of a high-dimensional physical system, such as the Franka robotic arm, using a Transformer-based architecture without prior knowledge of the system's physical parameters. The objective is to predict quantities of interest (end-effector pose and joint positions) given the torque signals for each joint. This prediction can be useful as a component for Deep Model Predictive Control frameworks in robotics. The meta-model establishes the correlation between torques and positions and predicts the output for the complete trajectory. This work provides empirical evidence of the efficacy of the in-context learning paradigm, suggesting future improvements in learning the dynamics of robotic systems without explicit knowledge of physical parameters. Code, videos, and supplementary materials can be found at \href{https://sites.google.com/view/robomorph}{project website}.
}

\onecolumn \maketitle \normalsize \setcounter{footnote}{0} \vfill

\section{Introduction}\label{sec:introduction}

The field of Deep Learning has undergone a significant transformation with the widespread adoption of Transformer-based architectures \cite{vaswani2017attention}, particularly impacting Natural Language Processing (NLP) in generating text \cite{touvron2023llama}. Recently, new applications have emerged to solve partial differential equations and image vision tasks. However, in complex areas such as robotics, \cite{shahid2022continuous}, where systems are highly non-linear, the implementation of Transformer-based solutions remains limited. Transformers have shown success in providing high-level task knowledge to robots, but there has been little progress in using them for learning system dynamics or performing system identification. This paper introduces a new approach to learning a meta-dynamical model for high-dimensional physical systems, such as the Franka robot arm, using a Transformer-based architecture without prior knowledge of the system's physical parameters. The goal is to accurately predict key quantities (end-effector pose and joint positions) based on the torque signals for each joint. Such predictions are valuable for integration into Deep Model Predictive Control frameworks, which are increasingly utilized in robotics. The meta-model is given an initial context, that establishes the relation between torques and positions and predicts the output for the complete trajectory. Using massively parallel simulations, large datasets, representing different robot dynamics, are generated in a simulated physics environment (Isaac Gym) to train the meta-model. The effectiveness of this learned model is demonstrated across various types of control inputs. This work demonstrates the use of transformer-based models in learning robot dynamics without any explicit knowledge of physical parameters.

\subsection{Related work}
In the domain of Partial Differential Equations (PDEs) resolution, Transformers are gaining momentum compared to Physics Informed Neural Networks (PINNs) and Recurrent structures such as LSTMs \cite{schmidhuber1997long}. Transformer-based architectures, coupled with Fourier-Neural-Operators \cite{li2020fourier}, demonstrate promising capabilities \cite{yang2023context}. Despite the limited existing literature on meta-learning in robotics, \cite{gupta2022metamorph} demonstrates the adaptability of Transformers in learning general controllers. Furthermore, Transformers and meta-learning have notably found applications in the domain of Fault Diagnosis \cite{10138459}. \cite{lin2022survey} discusses considerations regarding Transformers and highlights a key challenge: their inefficiency in processing long sequences, primarily due to the computational and memory complexities associated with the self-attention module.

\subsection{Contribution}
\label{sec:Problem_Statement}
This paper proposes an approach that utilizes transformer-based architectures to learn a meta-dynamical model of a robot arm. Diverse training datasets are generated using massively parallel simulations in the Isaac Gym simulation framework~\cite{DBLP:journals/corr/abs-2108-10470}. To generate trajectories for robots in simulation, specific families of Operational Space Control (OSC) tasks were selected, with domain randomization applied to certain parameters. After learning the meta-model, this work investigates the following aspects:
\begin{enumerate}
    \item \textbf{Hyper-parameters and context}: Evaluation of how hyper-parameters, such as the number of multi-attention heads, dimension of compressed information ($d_{\rm model}$), and context length influence prediction accuracy.
    \item \textbf{In-Distribution and Out-of-Distribution Performance}: Examining the meta-model's ability to generalize in an out-of-distribution regime, and the required training dataset dimension to achieve acceptable results.
    \item \textbf{Transfer Learning on Out-of-Distribution Scenarios}: Analysis of the impact of pre-training on the transfer of knowledge across different distributions of control actions, including zero-shot and few-shot learning scenarios.
\end{enumerate}

\section{Methods}
\label{sec:Background}
This paper aims to develop an approach that relies on a black-box model-free simulation of a system without any information regarding the robotic system.

\subsection{Preliminaries}

\textbf{Black Box Model}: Black box models describe the input-output behavior of a system without explicitly modeling its internal mechanisms. These models are estimated directly from experimental data.

\noindent \textbf{Simulation Model}: Simulation models predict system outputs solely from inputs, without relying on past measured data. They are valuable for designing controllers and emulating physical systems.

\noindent \textbf{Model-free simulation:} In this approach no specific model is needed for the system at hand; the meta-model $\free_{\phi}$ receives input/output sequences ($u_{1:m}^{(i)}, y_{1:m}^{(i)}$) up to time step $m$ (\emph{context}) 
and a test input sequence (\emph{query})  $u_{m+1:N}^{(i)}$ from time $m+1$ to $N$ to produce the corresponding output sequence 
$\hat y_{m+1:N}^{(i)}$: 
\begin{equation}
\label{eq:model_free_sim}
\hat y_{m:N}^{(i)} = \free_\phi(u_{1:m-1}^{(i)}, y_{1:m-1}^{(i)}, u_{m:N}^{(i)}).
\end{equation}
The meta-model indeed is trained by minimizing the mean squared error (MSE) between the simulated open-loop prediction and the ground truth.

\subsection{Model architecture}
The standard Transformer architecture has been adapted to handle real-valued input/output sequences generated by dynamical systems, instead of the sequences of symbols (word tokens) typically used in natural language modeling. Specifically, compared to plain GPT-2 \cite{radford2019language}, the initial token embedding layer is replaced by a linear layer with $n_u + n_y$ inputs and $d_{\rm model}$ outputs, while the final layer is replaced by a linear layer with $d_{\rm model}$ inputs and $ n_y $ outputs. This modification allows the Transformer to effectively process continuous-valued sequences, typical of dynamical systems, aligning the model architecture with the requirements of system identification tasks rather than natural language processing.

\begin{figure} 
\centering
\includegraphics[width=0.45\textwidth]{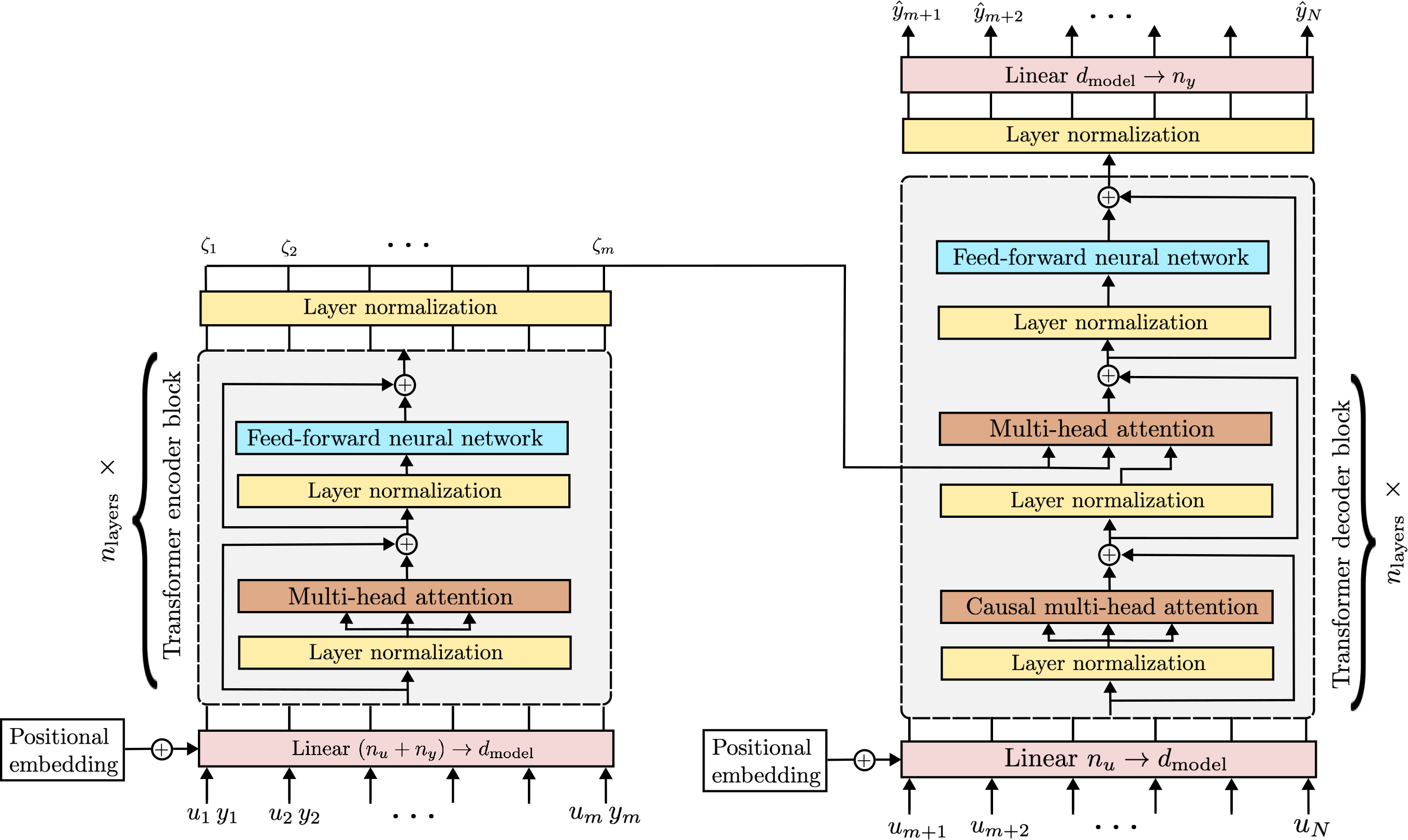}
 \caption{Encoder-decoder Architecture \cite{forgione2023from}.} \label{fig:Used model}
\end{figure}

The architecture is visualized in Figure \ref{fig:Used model} and consists of (i) an encoder that processes $ u_{1:m} $, $ y_{1:m}$ (without causality restriction)
and generates an embedding sequence $ \zeta_{1:m}  $; (ii) a decoder that processes $ \zeta_{1:m}$ and test input $ u_{ m+1:N }$ (the latter with causality restriction) to produce the sequence of
predictions $ \hat{y}_{m+1:N} $.


\section{Experimental framework}
\begin{figure}[b]
    \centering	
     \includegraphics[width=0.5\textwidth]{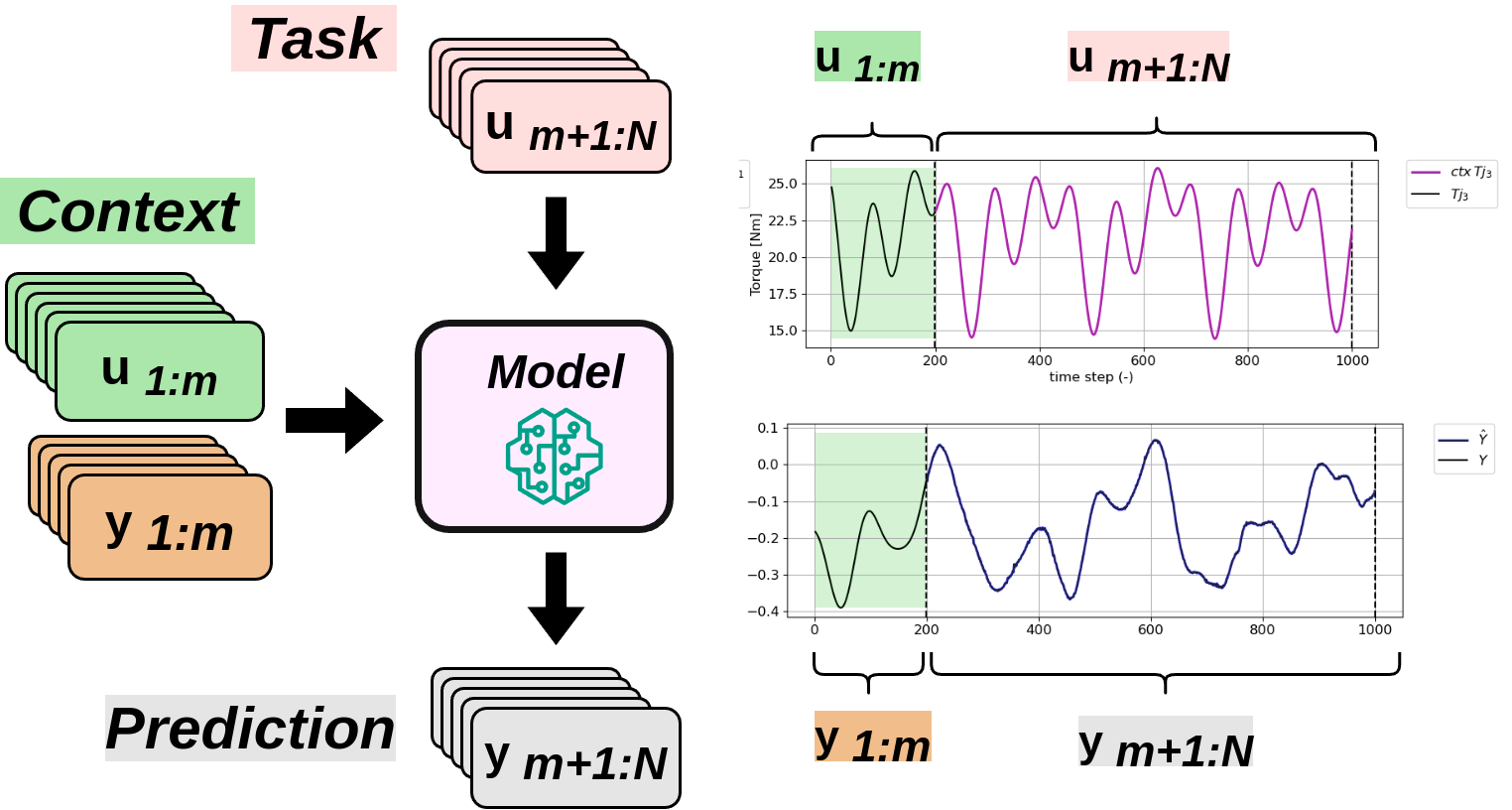}
     \vspace{2pt}
    \caption{Proposed meta-model uses context and input to perform the prediction. Context length is highlighted in green color.}
    \label{fig:test_pipeline}
\end{figure}

\subsection{Overview}
The strength of the proposed approach lies in its flexibility, allowing us to arbitrarily choose the dimensions of inputs and outputs. In the analyzed experimental framework, we learn a mapping from a 7-dimensional input to a 14-dimensional output:

\begin{itemize}
\item \textbf{Input}: joint torques applied to the 7 degrees of freedom (DoFs) Franka robot in $\boldsymbol{Nm}$.
\item \textbf{Output}:
    \begin{itemize}
        \item end effector (EE) Cartesian coordinates $(x,y,z)$ in [m];
        \item EE quaternion $(X,Y,Z,W)$;
        \item joint positions $(q_0 \dots q_6)$ in [rad].
    \end{itemize}
\item \textbf{Context}: 20\% of the entire simulation. 
\item \textbf{Total time window}: 1000 steps → 16.7 s.
\end{itemize}


The core idea is to train a meta-model by varying the range of parameters influencing the dynamics of the Franka robot. Figure \ref{fig:test_pipeline} shows the context and the input used to generate the predictions for complete trajectory.

\subsection{Domain randomization}
The classic domain randomization approach simulates a broad distribution across domains or Markov Decision Processes (MDPs), aiming to train a model robust enough to perform well in real-world conditions \cite{hospedales2020metalearningneuralnetworkssurvey}. Our proposed approach focuses on modeling a real Franka robot arm using this consideration: Handling real-world scenarios involves addressing deviations from nominal values, such as uncertainties in joint friction and damping characteristics.
Below, we highlight some foundational aspects of the proposed domain randomization:
\begin{itemize}
    \item Each link's mass is randomized according to a uniform distribution. This strategy increases the complexity of the problem, prevents overfitting in the model, and enhances overall robustness.
    \item Initial joint positions are randomized to facilitate meta-learning of the system. This variation aims to enhance the explorability of the robot's workspace, rather than strictly mapping specific control actions or trajectories of the robotic arm.
    \item Unlike typical robotic controllers in Franka Control Interface (FCI ), which compensates for gravity terms and internal joint frictions by default, our control actions consider those components. We virtually handle the torque from the motor side, emphasizing that the model needs to gain a deeper understanding of the underlying physical system.
\end{itemize}

\subsubsection{Object of randomization}
\textbf{Initial Joint Positions}: Initial joint positions are randomized around the mid-positions.\\
\textbf{Mass of the Links}: Each link's mass is uniformly distributed around its \textit{nominal value}, with a symmetric variation of a certain percentage ($\pm\,x\,$\%).\\
\textbf{Joint stiffnesses}: Stiffness and damping of joints are randomized and constrained to fixed values.\\
\textbf{Center of masses}: The center of mass is varied along three dimensions.\\
\textbf{Frequency (for random inputs)}: This parameter serves as the main frequency component of the overall signal. Its value ranges from 0.1 and 0.25 Hz.\\

\subsection{Task definition}
\begin{figure}[b!]
	\centering
	\includegraphics[width=0.4\textwidth]{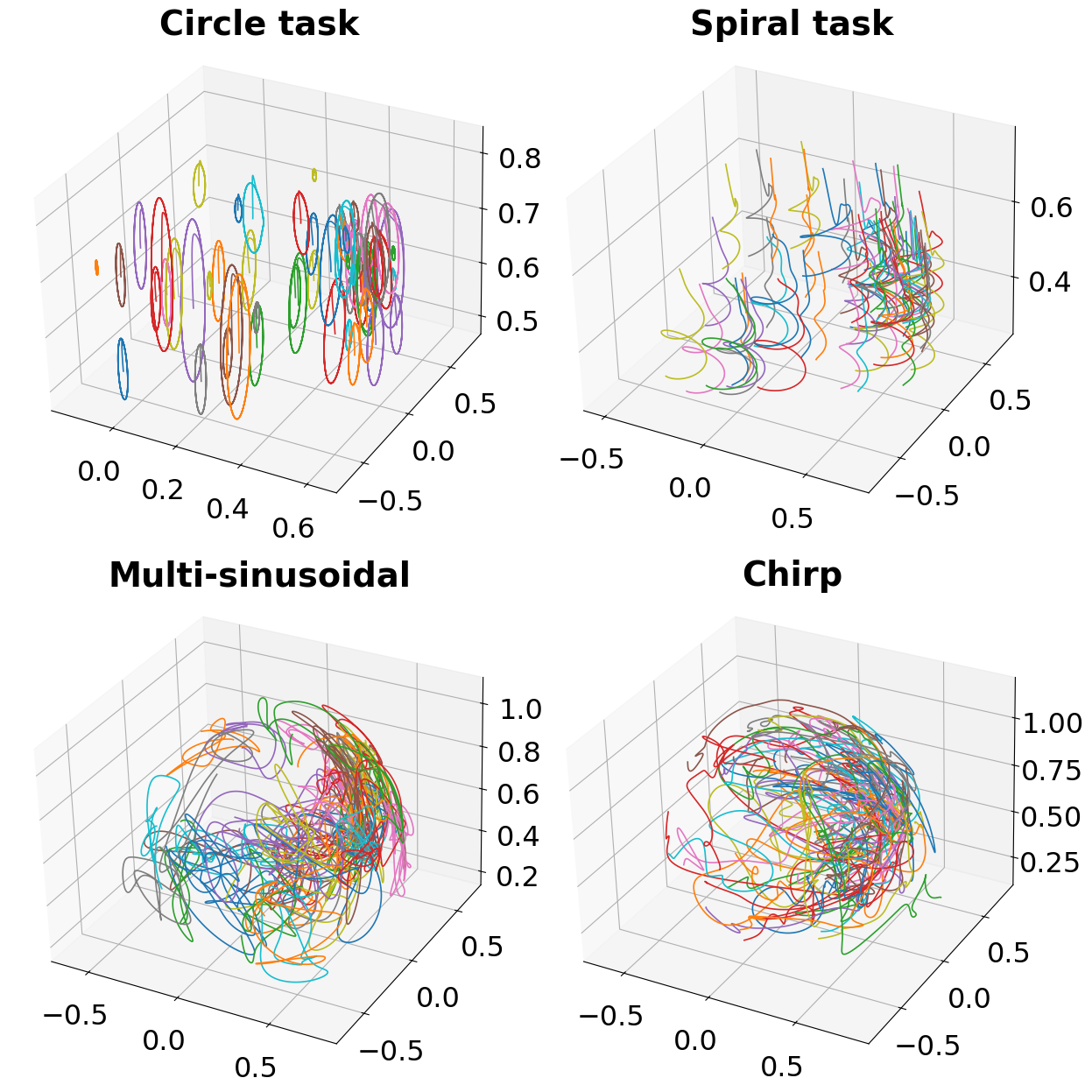}
	\caption{3D trajectories in Cartesian coordinates [m].}\label{fig:tasks}
\end{figure}

Tasks are categorized into two main families: synthetic inputs or trajectories derived from Operational Space Control (OSC), spanning from direct torque commands to desired Cartesian positions. Figure \ref{fig:tasks} shows the example cartesian trajectories generated with two types of control inputs.

\subsubsection{Synthetic random input}
In this case, the torque is generated directly by a specific mathematical function and applied in $\boldsymbol{Nm}$.
Two types of torque inputs, namely \textit{multi-sinusoidal} and \textit{chirp}, are considered to generate torque profiles for each joint:

\vspace{2pt}
\begin{itemize}
    \item \textit{Multi-sinusoidal:}
    \begin{footnotesize}
        \begin{equation}
            u_i = \begin{bmatrix} A_0 & A_1 & A_2 & A_3 \end{bmatrix}_i \begin{bmatrix} \cos(w_0 t) \\ \sin(w_1 t) \\ \cos(w_2 t) \\ \sin(w_3 t) \end{bmatrix}_i,
        \end{equation}
        \begin{align}
        \begin{bmatrix} w_0 \\ w_1 \\ w_2 \\ w_3 \end{bmatrix} &= \begin{bmatrix} w_0 \\ 1.5 \cdot w_0  \\ 2 \cdot w_0 \\ 3 \cdot w_0 \end{bmatrix},
         \\ w_0 = 2 \pi f_0, & \quad f_0 \in \left[ \frac{f_{m}}{1.5},1.5 \cdot f_{m}\right] ,
        \\ A_0 \cdots A_3 \in & \left[ -f_m \cdot 15 , f_m \cdot 15 \right].
        \end{align}
        \end{footnotesize}
    \item \textit{Chirp:} 
    \begin{footnotesize}
        \begin{equation}
            u_i = A_i \cos (\,w_1 \cdot(\,1 + \frac{1}{4} \cdot \cos(\,w_2 \cdot t)) \cdot t + \phi),
        \end{equation}
        \begin{align}
        \phi \in [\, - \pi , \pi \,], \quad q_0 \in [\, - 0.5 , 0.5\,],  \quad A_i \in [\, - 4 , 4 \,],
        \\ w_1 =2\pi f_1, \quad  w_2 = 2\pi f_2 ,
        \\  f_1 \in [\, f_m , f_m \cdot 1.5 \,], \quad f_2 \in [\, \frac{f_m}{1.5} ,  f_m \cdot 2 \,].
        \end{align}
    \end{footnotesize}
\end{itemize}
    
\begin{figure}[b]
	\centering
	\includegraphics[width=.4\textwidth]{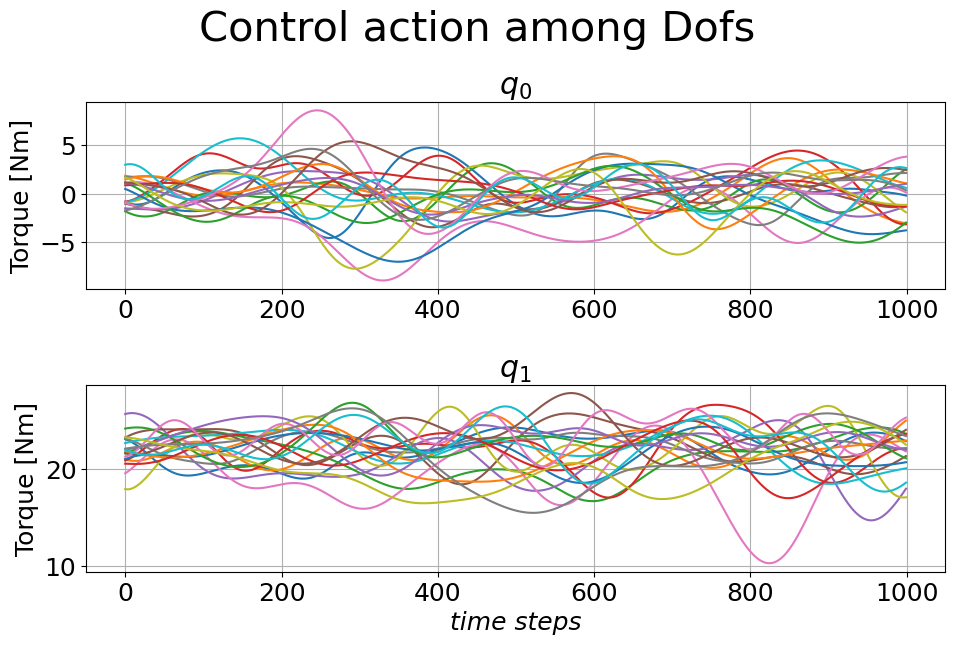}
	\caption{Multi-sinusoidal torque profiles for joint 0 and joint 1, respectively, and for 20 robots each.}
	\label{fig:Multisinusoidal}
\end{figure}

Figure \ref{fig:Multisinusoidal} illustrates random control actions for two joints generated with multi sinusoidal, with each color representing a different robot. It is important to note that joint 1 experiences higher gravity compensation compared to joint 0. As a result, the randomness appears less pronounced but it also varies considerably across the workspace.

\subsubsection{Operational space control (OSC)}
OSC tasks depend on two control gains, $\mathbf{K_p}$ and $\mathbf{K_d}$ which regulate the responsiveness of the control action in relation to the error with respect to the desired trajectory $\boldsymbol{x}^d$. The control input $\boldsymbol{u}$ is defined as:

\begin{align*}
    \boldsymbol{u} &= \boldsymbol{g}(\boldsymbol{q}) + \boldsymbol{J}_{\text{ee}}^{T}(q) \Big( \boldsymbol{M}_{\text{ee}} \cdot \boldsymbol{\ddot{x}}^d + \boldsymbol{C}_{\text{ee}} \cdot \boldsymbol{\dot{x}}^d \nonumber\\
    &\quad + \boldsymbol{K}_{p} \cdot (\boldsymbol{x}^d - \boldsymbol{x}) + \boldsymbol{K}_{d} \cdot (\boldsymbol{\dot{x}}^d - \boldsymbol{\dot{x}}) \Big) \nonumber.
\end{align*}

Two operational space tasks are defined as illustrated in Figure \ref{fig:tasks}:
    \vspace{1pt}
    \begin{itemize}
        \item \textit{Circle Task}: circle trajectories on YZ plane, with different radii and \textit{same} frequencies.
        \item \textit{Spiral Task}: spiral trajectories on the XY plane performed top-down or down-top, at different frequencies and with different radii.
    \end{itemize}

\subsection{Data generation in Isaac Gym}
\begin{figure}[t!]
	\centering
	\includegraphics[width=0.205\textwidth]{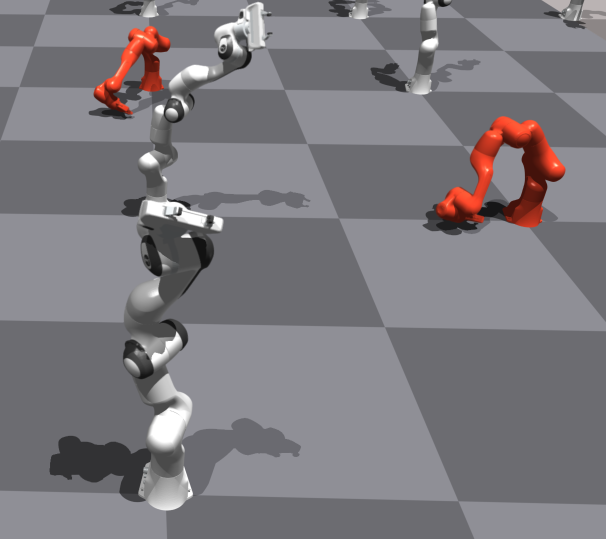}
        \includegraphics[width=0.15\textwidth]{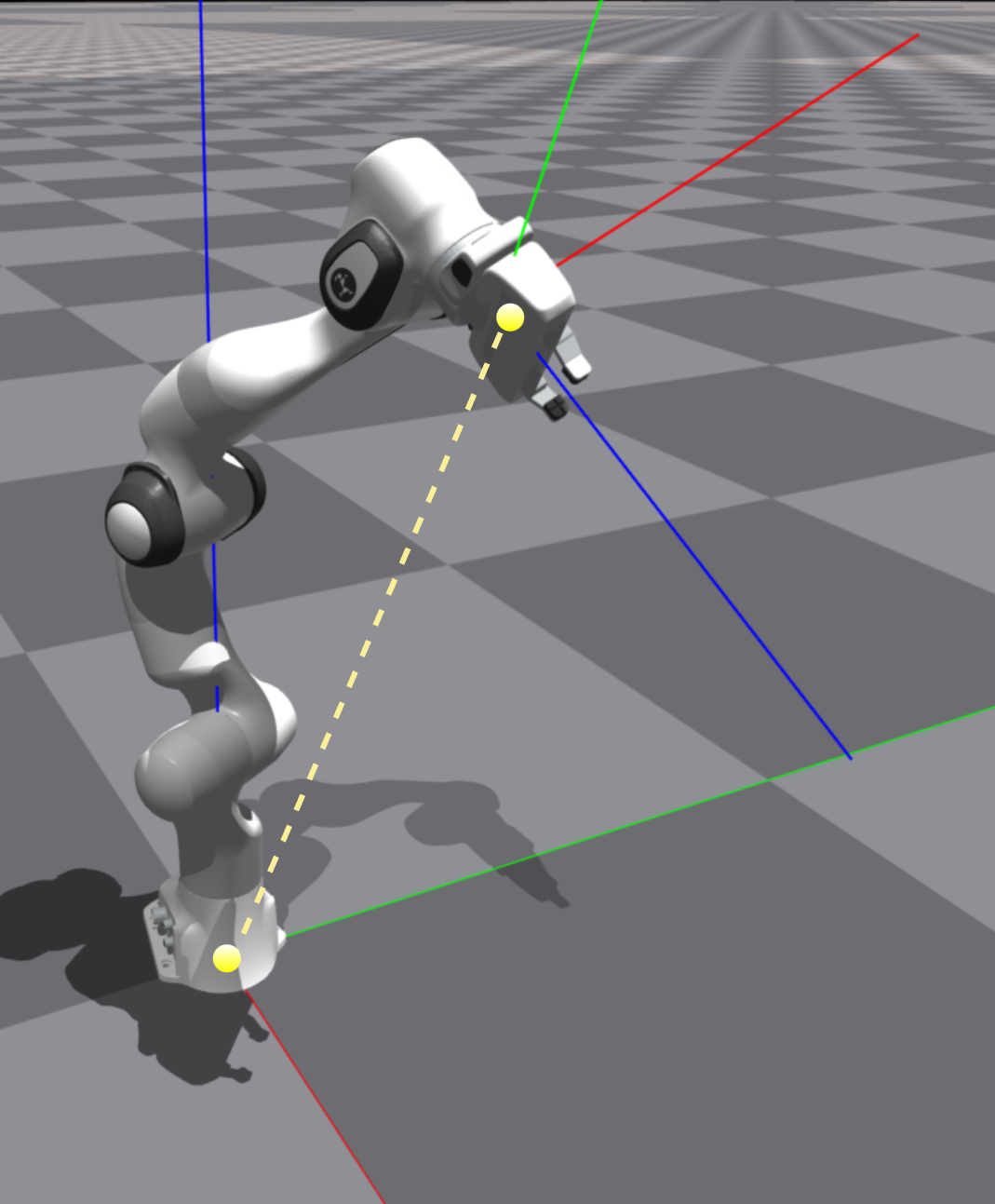}
	\caption{Collision detection visualization.}
	\label{fig:collision_detection}
\end{figure}

Isaac Gym \cite{DBLP:journals/corr/abs-2108-10470} is NVIDIA’s prototype physics simulation environment for RL research. It allows developers to experiment with end-to-end GPU-accelerated RL for physical systems. To the best of the author's knowledge, there is no previous contribution mentioning Isaac Gym as a simulation environment outside of the RL domain, indeed the use of Isaac Gym in this work to generate large datasets for transformer-based model training is quite novel.
Ensuring a trade-off between randomness and feasibility regarding workspace position and required torque is crucial to prevent issues during the training stage. For these reasons, three features/sub-functions have been incorporated to handle spurious robot instances:

\begin{itemize}
    \item \textbf{Self-collision and floor-collision detection}:
    To ensure self-collision and floor-collision scenarios are managed effectively, a net contact force tensor is employed to filter out colliding robots from the data buffer.
 Figure \ref{fig:collision_detection} shows the detection of such collisions.

    \item \textbf{Position and torque saturation check}: In this scenario training data should ideally consist of numerical functions that are free from singularities or unrealistically high values. At each time step, simulations that result in joint positions reaching their limits or torques which saturate, are excluded from the training dataset. 

    \item \textbf{Exclusion of quaternion error}: In certain simulations, unexpected changes in quaternion acquisition have been observed, possibly due to internal reference shifts within Isaac Gym or singularities. These sudden changes occurred infrequently and were straightforwardly excluded from the dataset. 
\end{itemize}

The data generation algorithm is shown in Algorithm \ref{algo:datagen}.

\begin{algorithm}[ht]
 \caption{Data Generation - Higher Level}\label{algo:datagen}
 \KwData{num\_robots, timesteps}
 \KwResult{Generated data, black-listed robots}
 
 Setting up simulation parameters\;
 Generating random torques for all robots' DoFs\;
 
 \For{$i \leftarrow 1$ \KwTo $num\_robots$}{
    Randomization of the dynamical parameters\;
 }
 
 \While{$t \leq timesteps$}{
    Apply torques and step simulation\;
    
    \eIf{$saturation$ \textbf{or} $collision$ \textbf{or} $self\_collision$}{
        add to the black-list of $robot\_idx$\;
    }{
        Store full pose and torques in buffer\;
    }
 }
 
 Remove black-listed robots\;
 Save tensors\;

\end{algorithm}

\subsection{Dataset composition}
\begin{figure}[H]
	\centering
	\includegraphics[width=0.33\textwidth]{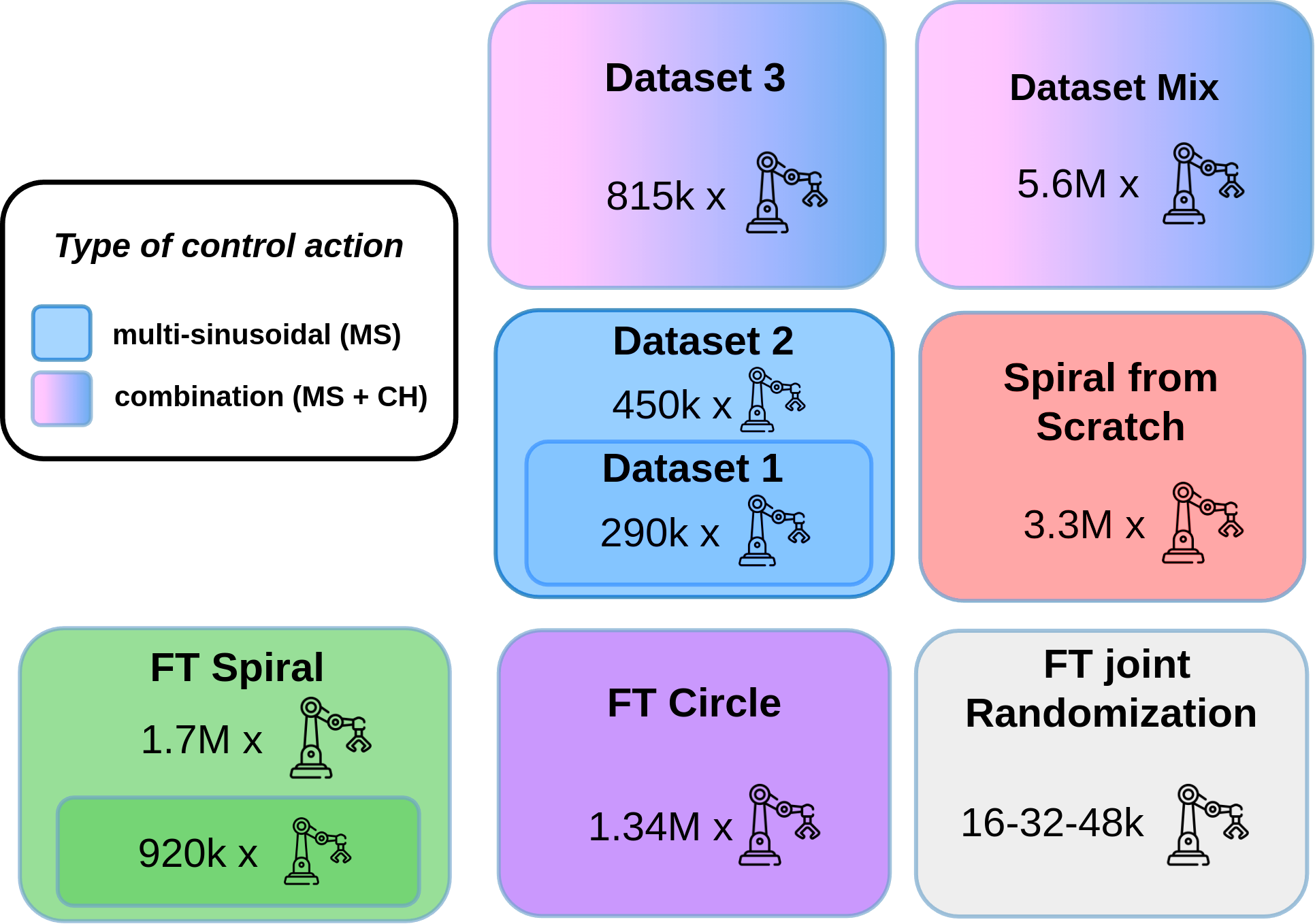}
	\caption{Dataset composition.}
	\label{fig:Dataset}
\end{figure}

Figure \ref{fig:Dataset} illustrates the composition of datasets used for training and fine-tuning based on the type of control action and the number of simulations. Each dataset has specific mass and joint randomization bounds. 

\subsection{Training loss}
\begin{figure}[b!]
	\centering
	\includegraphics[width=.45 \textwidth]{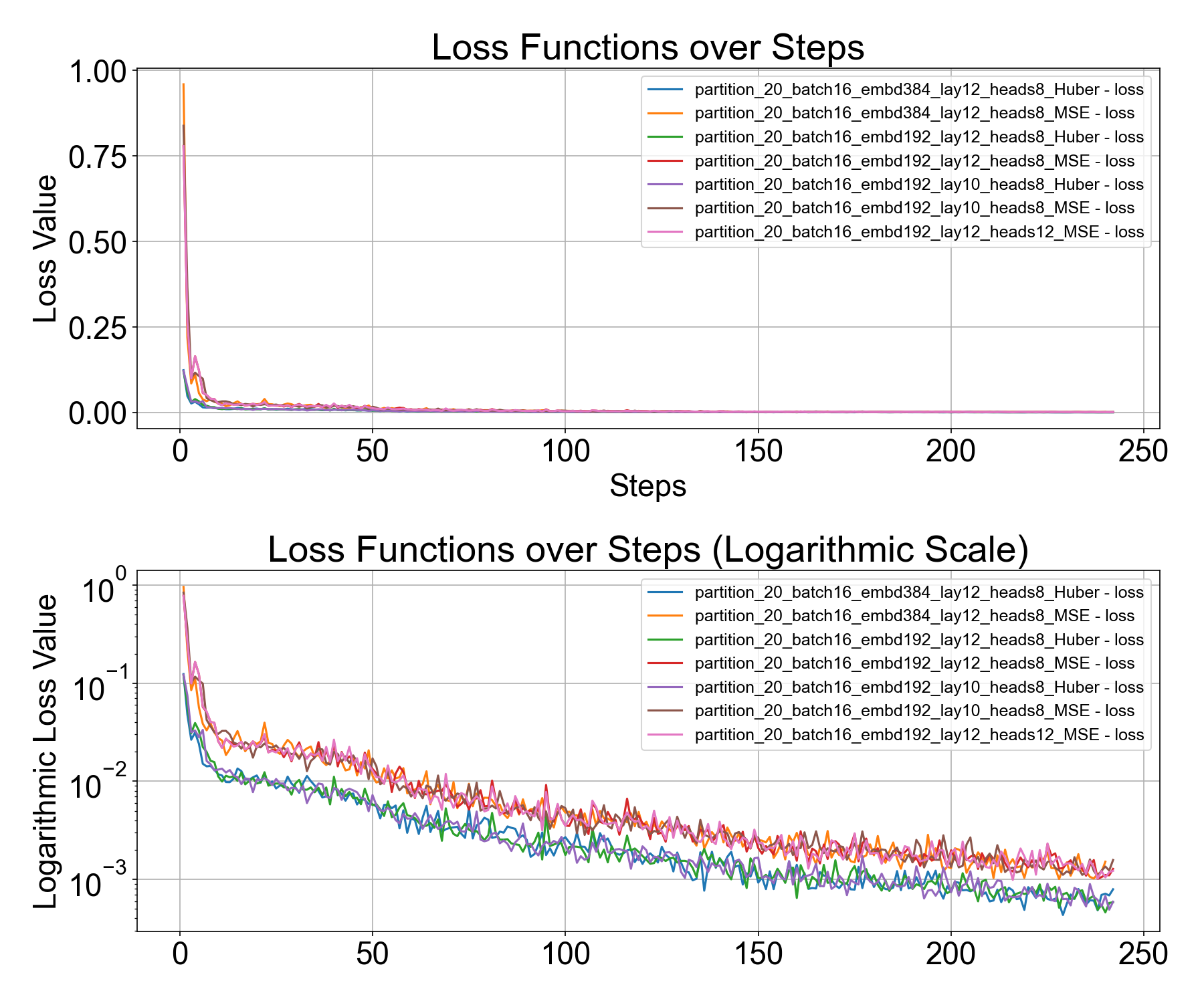}
	\caption{Loss function of an example dataset (for every 100 iterations).}
	\label{fig:loss_trend}
\end{figure}

As highlighted by \cite{kirsch2024generalpurpose}, Transformer training often exhibits a substantial plateau problem. One of the solutions proposed by the authors is to increase the batch size to reduce the plateau problem. In this work, each ``simulation batch'' is composed of over 3000 classes of Franka robots (each with different dynamics), while each training batch consisted of 16 robots. This approach aimed to promote learning-to-learn as opposed to mere memorization. The loss function for an example training is shown in Figure \ref{fig:loss_trend}.

\section{Evaluation approach}
Each model trained on a specific dataset is tested on different simulations, which may be In or Out of Distribution with respect to the training dataset. Different metrics have been used to compare these prediction performances.

\label{sec:Model_Evaluation}

\subsection{Metrics}
\begin{figure}[t]
	\centering
	\includegraphics[width=0.4\textwidth]{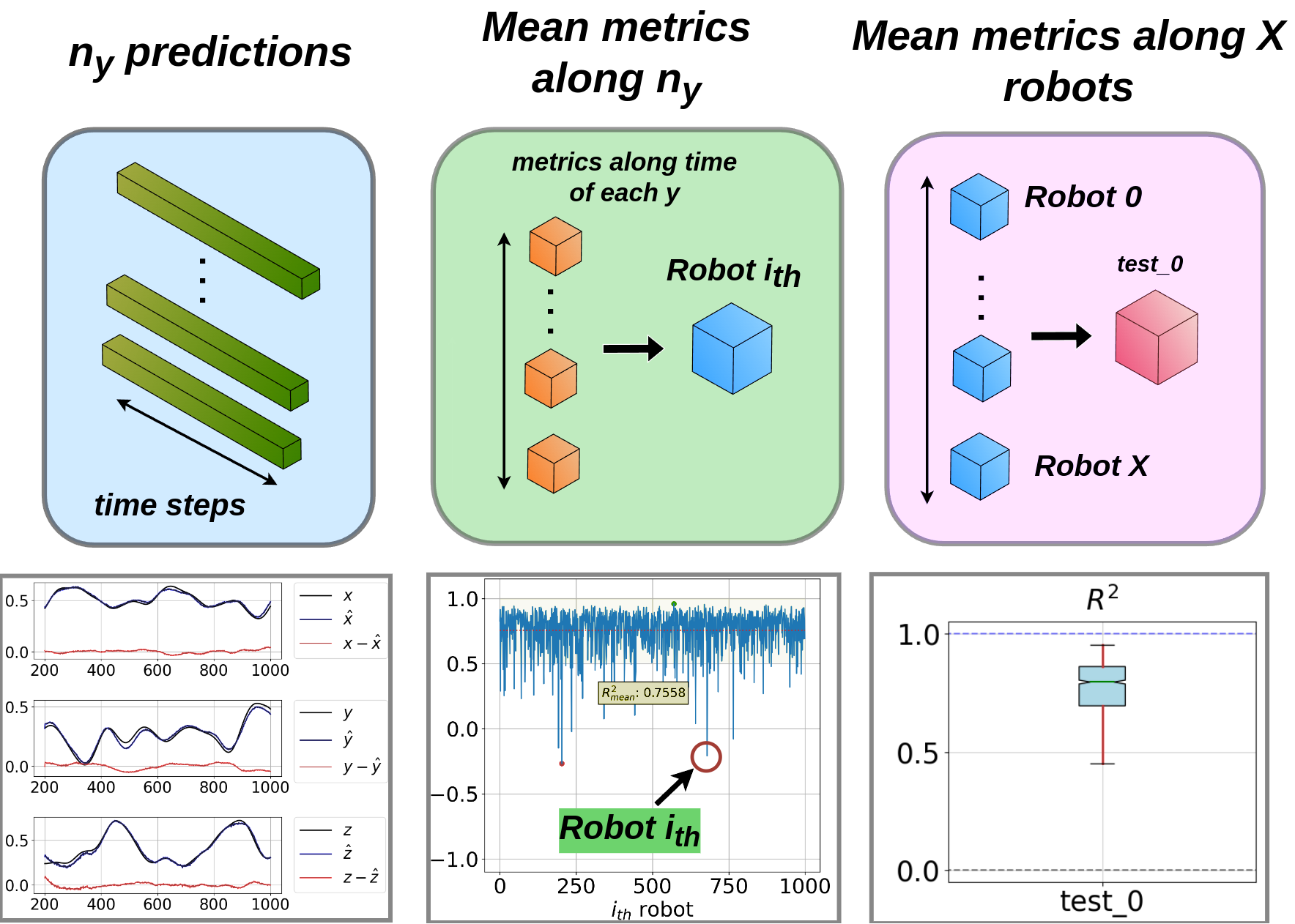}
	\caption{First evaluation approach.}
	\label{fig:test}
\end{figure}

Indexes are calculated over time for each coordinate, then averaged across the output dimensions $n_y$, and finally averaged across a batch of $X$ robots. This type of evaluation is illustrated in Figure \ref{fig:test}. The following indexes are used:
\begin{itemize}
    \item Coefficient of Determination:
    $R^2= 1 - \frac{\sum_{i=1}^{n} (y_i - \hat{y}_i)^2}{\sum_{i=1}^{n} (y_i - \bar{y})^2}$;
    \item Root Mean Square Error: $\rm{RMSE} = \sqrt{\frac{1}{n} \sum_{i=1}^{n} (\hat{y}_i - y_{i})^2}$;
    \item Normalized Root Mean Square Error: $\rm{NRMSE} = \frac{\text{RMSE}}{\sigma_{y}}$;
    \item Fit Index: $FI = 100 \cdot \left(1 - \frac{\sqrt{ \norm {\sum_{i=1}^{n} (y(t) - \hat{y}(t))}^2}}{\sqrt{\sum_{i=1}^{n} (y(t) - \bar{y}(t))^2}}\right)$.
\end{itemize}


The main drawback of this approach is that $R^2$ for ``flat'' and relatively small values tends to penalize the single robot prediction metrics. In the OSC tasks, most of the joints experience small variations compared to multi-sinusoidal and chirp signals and for this reason, averaging over the output dimensions $n_y$ and then over batches leads to really low $R^2$ mean values. 
For this reason another {\textit{evaluation approach} was used to investigate the performance of OSC tasks. Instead of averaging over output dimensions $n_y$ and then over different robots, several trajectories are merged sequentially coordinate by coordinate, and then indexes are computed on these merged trajectories as shown in Figure \ref{fig:new_metrics}. Predictions are performed separately but they're considered merged only for the calculation of metrics. The aim is to better demonstrate the overall capability in different scenarios rather than evaluating the local precision as in the first evaluation approach.

\begin{figure}[t!]
	\centering
	\includegraphics[width=0.35\textwidth]{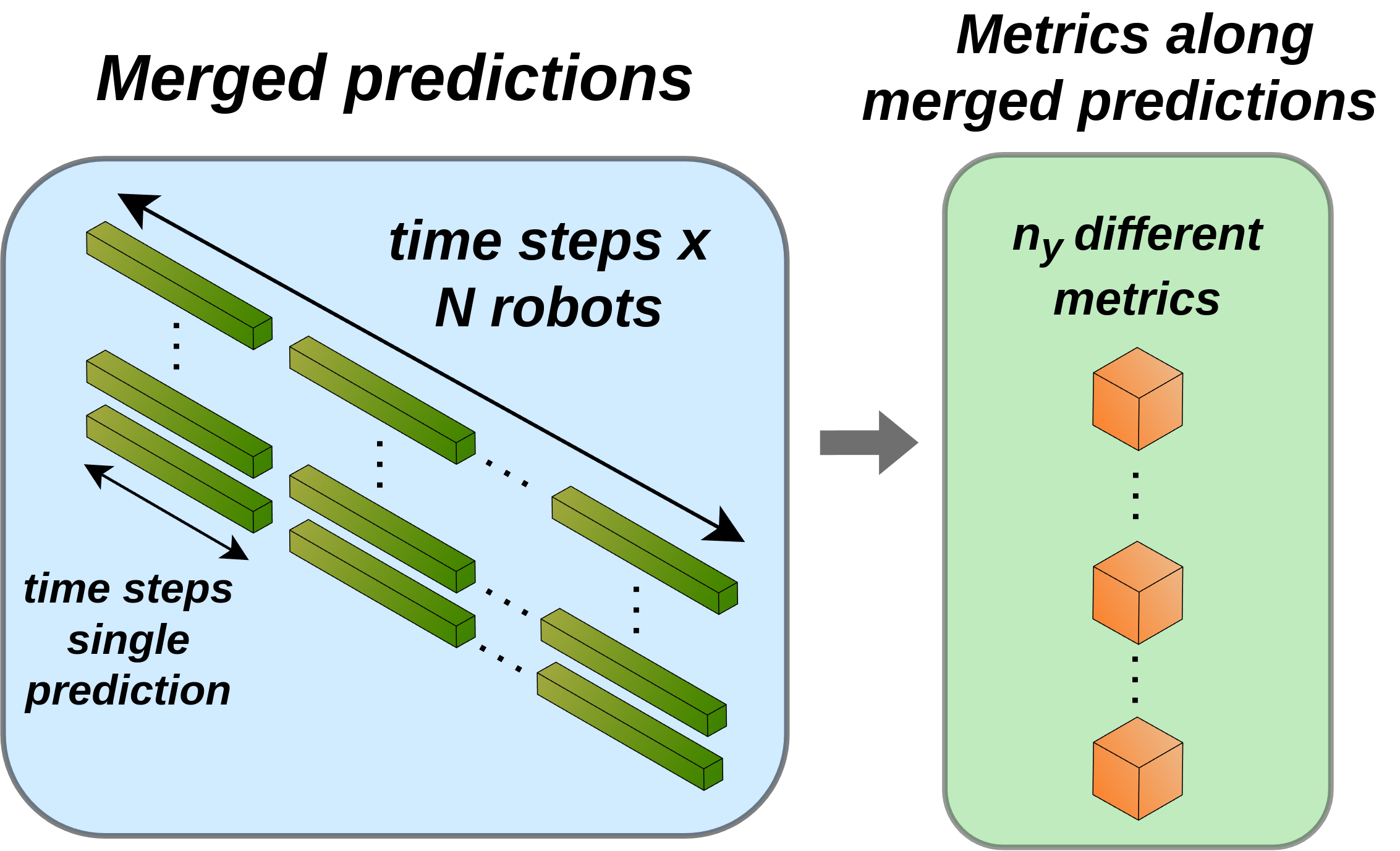}
	\caption{Second evaluation approach.}
	\label{fig:new_metrics}
\end{figure}

$R^2$ tends to assume values very close to 1.0 on the coordinates which show already good capabilities locally. To better evaluate the prediction performance of the models, the fit index has been chosen as the primary metric, due to its better differentiation even when $R^2$ values are particularly high and similar.

\section{Results}

\textbf{Influence of the training context:} Tests were performed for contexts ranging from 5\% to 50\%, with 20\% used as a reference for all tests. The quadratic dependence of computational and memory complexity on sequence length in self-attention is a limiting factor for Transformers. Generally, Transformers can accept context lengths different from those used during training, provided that the test context lengths are smaller than the training ones. Accordingly, it is also possible to use the same context length while varying the prediction horizon to a smaller number of steps, effectively changing the context length in percentage.

\begin{figure}[b!]
    \centering
    \includegraphics[width=.49\textwidth]{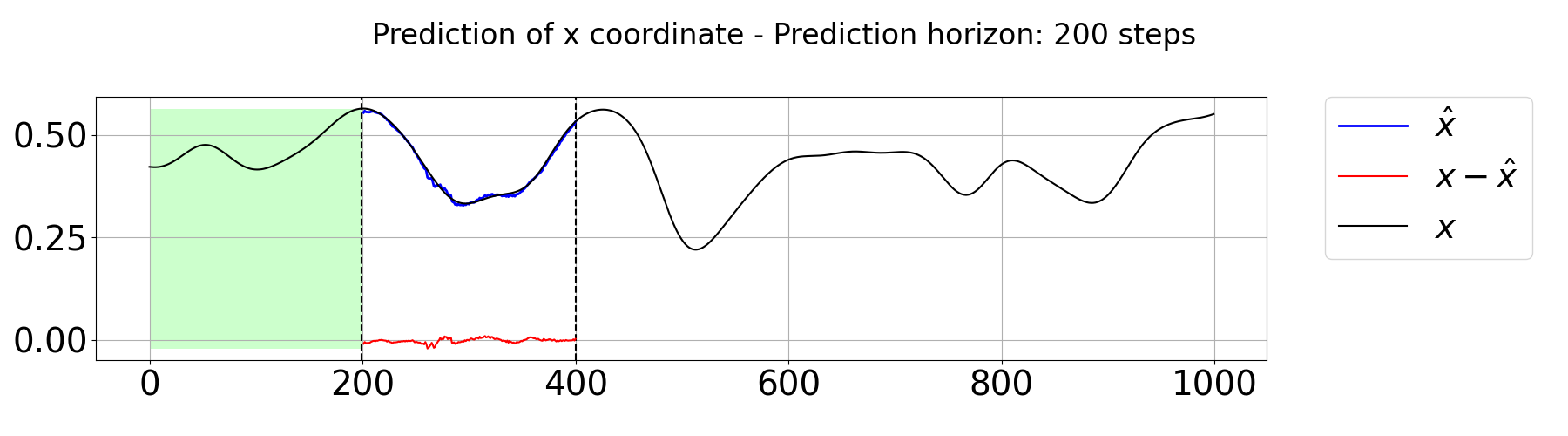}
    \includegraphics[width=.49\textwidth]{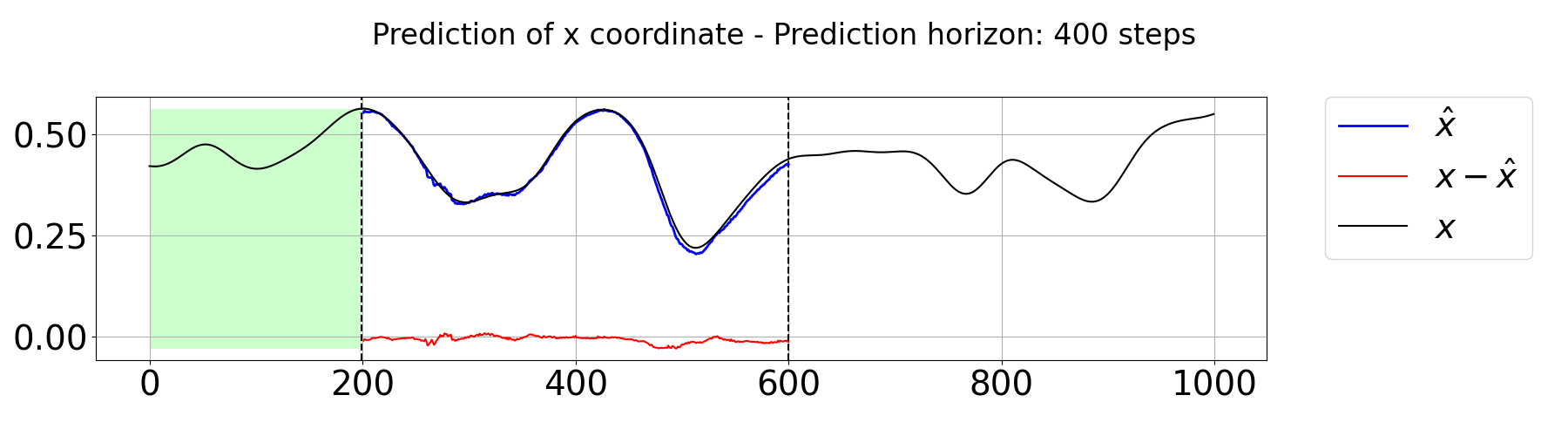}
    \includegraphics[width=.49\textwidth]{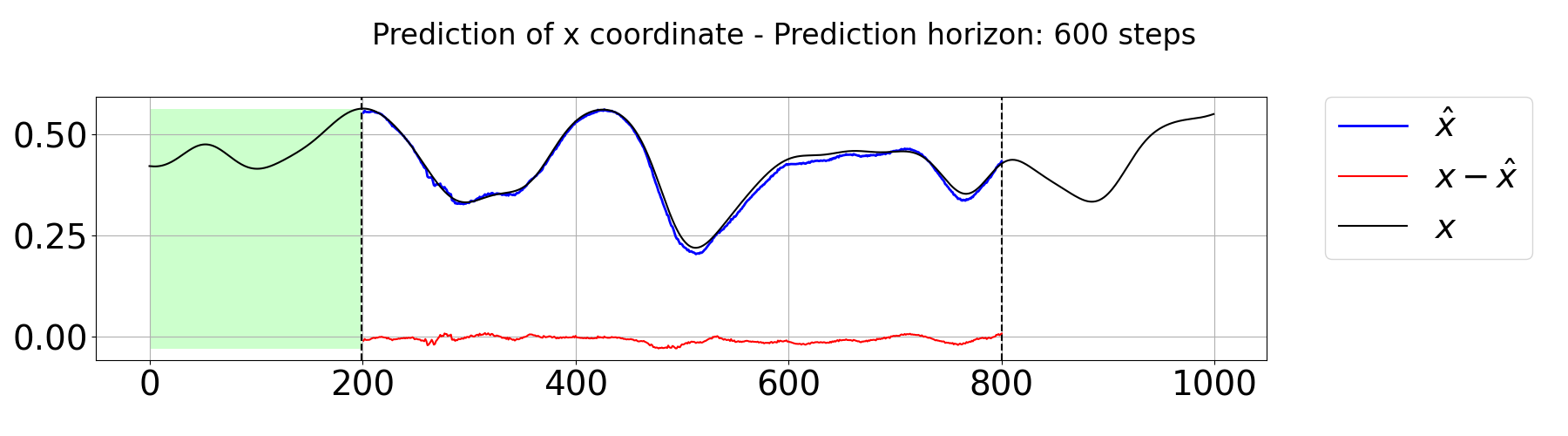}
    \caption{Different prediction horizons and with the same test context as in training.}
    \label{fig:different Horizons}
\end{figure}

Varying the prediction horizon while keeping the context window the same as in training reveals an initial transition error. However, overall performance improves as the test horizon approaches the training horizon as shown in Figures \ref{fig:different Horizons} and \ref{fig:Comparison}, indicating that the model requires the knowledge of whole trajectory patterns to perform optimally. 

\begin{figure}[t!]
    \centering
    \includegraphics[width=.4\textwidth]{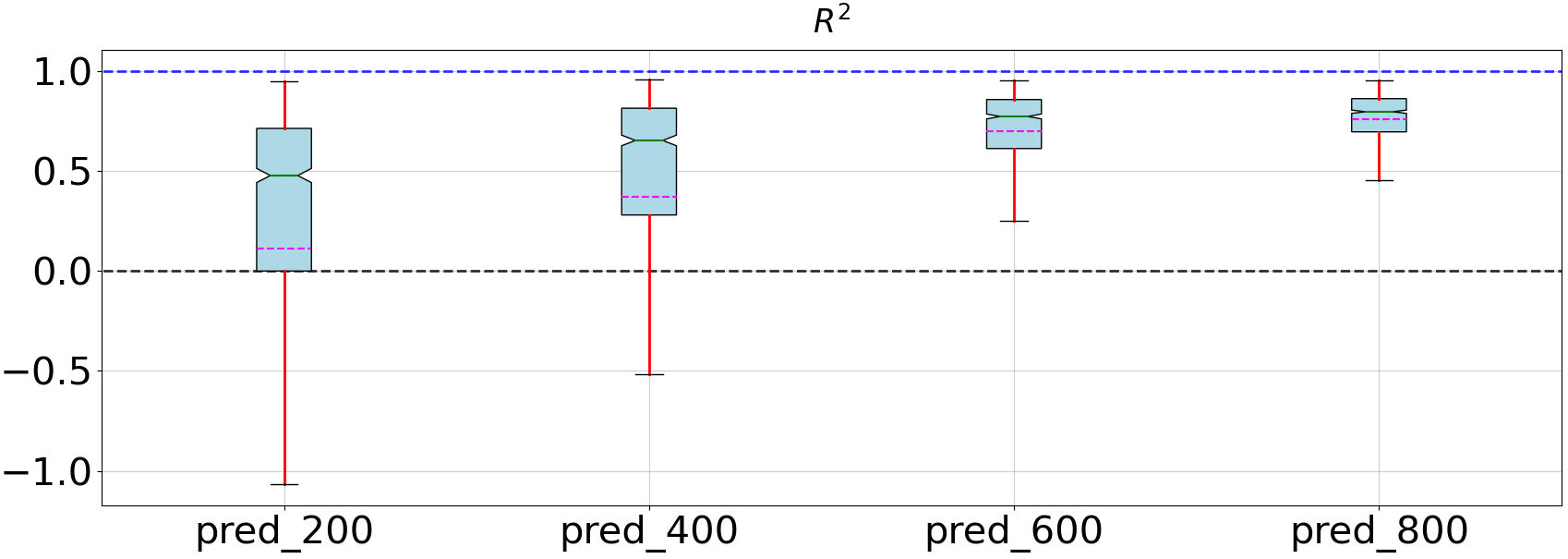}
    \caption{$R^2$ in a 200 steps context.}
    \label{fig:Comparison}
\end{figure}

\noindent \textbf{Hyper-parameters tests:} 
    The hyper-parameter values tuned through trial and error are given below. Most other parameters were fixed as per \cite{forgione2023from}. Batch size is not mentioned as an independent parameter, due to computational constraints. 
\begin{itemize}
	\item \textbf{Loss function}: MSE / Huber;
	\item \textbf{Model Dimension} (\textbf{$d_{\rm model}$}): 192 / 384;
	\item \textbf{Number of multi-attention heads}: 8 / 12;
	\item \textbf{Number layers}: 10 / 12 / 16.
\end{itemize}
\vspace{3pt}
Given the training and test dataset from the same distribution, it is possible to investigate how these parameters affect the quality of predictions. Table \ref{tab:layer_comp} shows an example demonstrating the impact of the number of layers on performance, using the first evaluation approach. The results highlight the best performance achieved with the 12 layers model. Varying the model dimension and number of multi-attention heads showed comparable results without any significant performance differences. The results are available in the appendix on project website.

\begin{table}[!b]
\centering
\scriptsize
\begin{tabular}{c|ccccc}
  \hline
\multicolumn{5}{c}{\textbf{10 Layers}} \\
  \hline
\textbf{TestA} & $R^2$ & $\sigma_{R^2}$ & RMSE & $\sigma_{RMSE}$ \\
  \hline
\textbf{test0} & 0.682 & 0.343 & 0.0530 & 0.0236 \\
\textbf{test1} & 0.750 & 0.292 & 0.0465 & 0.0218 \\
\textbf{test2} & 0.755 & 0.293 & 0.0458 & 0.0212 \\
  \hline
\multicolumn{5}{c}{\textbf{12 Layers}} \\
  \hline
\textbf{TestA} & $R^2$ & $\sigma_{R^2}$ & RMSE & $\sigma_{RMSE}$ \\
  \hline
\textbf{test0} & 0.713 & 0.287 & 0.0521 & 0.0242 \\
\textbf{test1} & 0.787 & 0.219 & 0.0454 & 0.0219 \\
\textbf{test2} & 0.793 & 0.212 & 0.0448 & 0.0213 \\
  \hline
  \multicolumn{5}{c}{\textbf{16 Layers}} \\
  \hline
\textbf{TestA} & $R^2$ & $\sigma_{R^2}$ & RMSE & $\sigma_{RMSE}$ \\
  \hline
\textbf{test0} & 0.700 & 0.303 & 0.0531 & 0.0258 \\
\textbf{test1} & 0.761 & 0.262 & 0.0468 & 0.0239 \\
\textbf{test2} & 0.767 & 0.255 & 0.0462 & 0.0234 \\
  \hline
  
\end{tabular}
\caption{Impact of the number of layers on prediction accuracies.}
\label{tab:layer_comp}
\end{table}
\noindent \textbf{Analysis of model performance:} The performance analysis of the optimal model is presented in Table \ref{tab:comparison} (with related color-convention in Table \ref{tab:convenzione}), with further details provided below:
\begin{itemize}
\item \textbf{In distribution}: For bounded tasks, good results can be achieved on in-distribution tests within 2 hours. However, as the number of tasks and frequencies increases, the required training time and model dimensions naturally increase.
\item \textbf{Slightly out of distribution}: The model showed good results, supporting the hypothesis of generalization rather than mere memorization of the training dataset. Specifically regarding mass randomization, its impact appears to be secondary to the frequency of the tested signal. Transformers demonstrate greater effectiveness when applied to higher frequencies compared to their training frequencies, as opposed to lower frequencies.
\item \textbf{Out of distribution}: The first column of Table \ref{tab:comparison} displays the performance of the model, which was trained on a mixed dataset containing both multi-sinusoidal and chirp signals, and tested on a spiral task. The performance varies significantly across different output dimensions, with some showing good results and others performing poorly.
\item \textbf{Fine-tuning on variable spiral}: Subsequent examinations involved a model trained extensively for 16 hours on a range of different frequencies, including both multi-sinusoidal and chirp signals. This model was then fine-tuned on a Spiral task for 2.5 hours (following the ideas in \cite{PiFoPuSYSID2024}). Results were compared to a model trained from scratch for 9 hours (Table \ref{tab:comparison}, third column). The results show that fine-tuned model perform significantly better than the one trained from scratch.
\end{itemize}

\begin{table}[t!]
\tiny
\centering
\begin{tabular}{c|cc|cc|cc}
  \hline
& \multicolumn{2}{c}{Zero-Shot} \vline & \multicolumn{2}{c}{FT Spiral (2.5h)} \vline & \multicolumn{2}{c}{Scratch (9h)} \\
  \hline
& $R^2$ &$Fit\,Index$ & $R^2$ & $Fit\,Index$ & R$^2$ & $Fit\,Index$ \\
  \hline
$x$ & \cellcolor{lightgreen!50}0.98& \cellcolor{lightgreen!50}86 & \cellcolor{lightgreen!80}0.998 & \cellcolor{lightgreen!80}96.09 & \cellcolor{lightgreen!50}0.988 & \cellcolor{lightgreen!50}88.92 \\
$y$ & \cellcolor{lightgreen!50}0.99 & \cellcolor{lightgreen!50}89.78 & \cellcolor{lightgreen!80}1 & \cellcolor{lightgreen!80}97.89 & \cellcolor{lightgreen!80}0.996 & \cellcolor{lightgreen!80}93.47  \\
$z$ & \cellcolor{red!30}-3.014 & \cellcolor{red!30}-100.34 &  \cellcolor{yellow!50}0.867 &  \cellcolor{yellow!50}63.57 & \cellcolor{red!30}-1.254  &  \cellcolor{red!30}-50.13  \\
$X$ & \cellcolor{lightgreen!50}0.986  & \cellcolor{lightgreen!50}88.01 & \cellcolor{lightgreen!80}0.999 & \cellcolor{lightgreen!80}97.56 & \cellcolor{lightgreen!50}0.982  & \cellcolor{lightgreen!50}86.54 \\
$Y$ & \cellcolor{lightgreen!50}0.989 & \cellcolor{lightgreen!50}89.51 & \cellcolor{lightgreen!80}1 & \cellcolor{lightgreen!80}98.04 & \cellcolor{lightgreen!80}0.998 & \cellcolor{lightgreen!80}95.72 \\
$Z$ & \cellcolor{red!30}0.137 & \cellcolor{red!30}7.11 & \cellcolor{yellow!50}0.953 &  \cellcolor{yellow!50}78.37 & \cellcolor{red!30}0.493 &\cellcolor{red!30}28.76 \\
$W$ & \cellcolor{red!30}0.191 & \cellcolor{red!30}10.05 &  \cellcolor{lightgreen!50}0.965 &  \cellcolor{lightgreen!50}81.39     & \cellcolor{orange!50}0.572   & \cellcolor{orange!50}34.54 \\
$q_0$ & \cellcolor{lightgreen!50}0.999  & \cellcolor{lightgreen!50}96.77 &  \cellcolor{lightgreen!80}1 &  \cellcolor{lightgreen!80}98.71   & \cellcolor{lightgreen!80}0.998 & \cellcolor{lightgreen!80}95.15 \\
$q_1$ & \cellcolor{lightgreen!50}0.966 & \cellcolor{lightgreen!50}81.56 &  \cellcolor{lightgreen!50}0.983  & \cellcolor{lightgreen!50}86.8   & \cellcolor{yellow!50}0.875 & \cellcolor{yellow!50}64.67 \\
$q_2$ & \cellcolor{yellow!50}0.847  & \cellcolor{yellow!50}60.94 &  \cellcolor{lightgreen!80}0.997  & \cellcolor{lightgreen!80}94.55  & \cellcolor{lightgreen!80}0.991 & \cellcolor{lightgreen!80}90.37 \\
$q_3$ & \cellcolor{red!30}-0.669  & \cellcolor{red!30}-29.18 &  \cellcolor{lightgreen!50}0.969 & \cellcolor{lightgreen!50}82.44  & \cellcolor{orange!50}0.627  & \cellcolor{orange!50}38.9 \\
$q_4$ & \cellcolor{lightgreen!50}0.988  & \cellcolor{lightgreen!50}88.96 &  \cellcolor{lightgreen!80}1 & \cellcolor{lightgreen!80}97.87   & \cellcolor{lightgreen!80}0.996 & \cellcolor{lightgreen!80}93.44 \\
$q_5$ & \cellcolor{yellow!50}0.871 & \cellcolor{yellow!50}64.13 &  \cellcolor{lightgreen!80}0.999 & \cellcolor{lightgreen!80}96.77 & \cellcolor{orange!50}0.838 & \cellcolor{orange!50}59.81 \\
$q_6$ & \cellcolor{lightgreen!50}0.981 & \cellcolor{lightgreen!50}86.11 &    \cellcolor{lightgreen!80}1 & \cellcolor{lightgreen!80}98.05 & \cellcolor{lightgreen!80}0.997  & \cellcolor{lightgreen!80}94.28 \\
  \hline
\end{tabular}
\caption{Comparison between zero-shot, fine-tuning on the spiral, and training from scratch.}
\label{tab:comparison}
\end{table}
\begin{table}[t!]
\scriptsize
\centering
    \begin{tabular}{c}
      \hline
        Fit Index Range \\
          \hline
         \cellcolor{lightgreen!80} $\geq$ 90 \\
         \cellcolor{lightgreen!50} 80 - 89.99 \\
         \cellcolor{yellow!50} 60 - 79.99 \\
         \cellcolor{orange!50} 30 - 59.99 \\
        \cellcolor{red!30} $<$ 30 \\
          \hline
    \end{tabular}
    \caption{Fit index color convention.}\label{tab:convenzione}

\end{table}

\noindent \textbf{Additional Analysis:}
Even after fine-tuning the Spiral task, challenges persist in the z-coordinate, likely due to the intrinsic complexity of the problem. Spirals, varying in both upward and downward directions, exhibit distinct characteristics in terms of frequencies and elevation rates: combinations of larger radii and slow elevations differ significantly from tasks involving fast elevation changes and small radii.
    
Fine-tuning on the same task family (multi-sinusoidal) with controlled joint randomization led to a significant increase in overall prediction accuracy, depending on the number of examples provided.  

Fine-tuning on Spiral tasks and zero-shot testing on the Circle task resulted in poor performance. This underscores the difficulty in predicting OSC tasks generally, particularly during initial transient phases. Direct fine-tuning on the Circle task yielded better results compared to fine-tuning on Spiral tasks and testing on Spiral. This observation suggests that the model's effectiveness is constrained to in-distribution (ID) and slightly out-of-distribution (OOD) tasks.

\section{\uppercase{Conclusions}}
\label{sec:conclusion}

This work tackles the problem of learning a meta-model of robot dynamics using an encoder-decoder Transformer architecture. The challenge lies in the simulation domain, where the meta-model accurately predicts complex systems over long sequences based on a 20\% context and 80\% prediction of the overall trajectory. The results indicate that Transformer-based models can learn dynamics in a zero-shot or few-shot fashion within control action distributions, suggesting their potential for use in robotics.
The results also highlight fine-tuning is advantageous in these scenarios and more practical compared to training such models from scratch.

Current limitations are mainly related to generalization. Following a black-box approach to generalize a single robotic arm independently of the type of control action appears structurally unfeasible. Results show distinctions between in-distribution (ID) and out-of-distribution (OOD) control actions, highlighting the critical role of the model's inputs. 

For future work, the results presented pave the way for pre-compensating control actions in unknown systems, particularly where estimating parameters such as payload, joint stiffness, and damping is challenging. This approach can be extended beyond control distributions to encompass Transfer Learning across diverse robot morphologies.

\section*{Acknowledgments}

This paper has received funding from the Hasler Foundation under the GENERAI (GENerative Robotics AI) Project.

\bibliographystyle{apalike}
{\small
\bibliography{robomorph}}

\end{document}